\documentclass[sigconf]{acmart}
\usepackage{graphicx}
\usepackage{algorithm}
\usepackage[noend]{algpseudocode}
\usepackage{amsmath}
\usepackage{caption}
\usepackage{subcaption}
\usepackage[group-separator={,}]{siunitx}

\def\BibTeX{{\rm B\kern-.05em{\sc i\kern-.025em b}\kern-.08emT\kern-.1667em\lower.7ex\hbox{E}\kern-.125emX}}
    
\copyrightyear{}
\acmYear{}
\setcopyright{none}
\acmConference[KDD-ADF-2019]{2nd KDD Workshop on Anomaly Detection in Finance}{August 5, 2019}{Anchorage, Alaska, USA}
\acmBooktitle{}
\acmPrice{}
\acmDOI{}
\acmISBN{}
\acmISBN{}

\begin{document}

%
\title{Automatic Model Monitoring for Data Streams}

%

\author{F\'abio Pinto}
\authornotemark[1]
\email{fabio.pinto@feedzai.com}
\affiliation{%
  \institution{Feedzai}
}

\author{Marco O. P. Sampaio}
\email{marco.sampaio@feedzai.com}
\authornote{Both authors contributed equally to this research.}

\affiliation{%
  \institution{Feedzai}
}

\author{Pedro Bizarro}
\email{pedro.bizarro@feedzai.com}
\affiliation{%
  \institution{Feedzai}
}

%

%

%
\begin{abstract}

	Detecting concept drift is a well known problem that affects production systems. However, two important issues that are frequently not addressed in the literature are 1) the detection of drift when the labels are not immediately available; and 2) the automatic generation of explanations to identify possible causes for the drift.
	For example, a fraud detection model in online payments could show a drift due to a hot sale item (with an increase in false positives) or due to a true fraud attack (with an increase in false negatives) before labels are available.
	In this paper we propose SAMM, an automatic model monitoring system for data streams. SAMM detects concept drift using a time and space efficient unsupervised streaming algorithm and it generates alarm reports with a summary of the events and features that are important to explain it.
	SAMM was evaluated in five real world fraud detection datasets, each spanning periods up to eight months and totalling more than 22 million online transactions. We evaluated SAMM using human feedback from domain experts, by sending them 100 reports generated by the system. Our results show that SAMM is able to detect anomalous events in a model life cycle that are considered useful by the domain experts. Given these results, SAMM will be rolled out in a next version of Feedzai's Fraud Detection solution.

\end{abstract}

%
%

%

%
\maketitle

\section{Introduction}

When a Machine Learning (ML) model is deployed into production in a data streaming scenario, the monitoring process that follows it is essential for the success of the Data Science project. The very nature of most data streams implies that they change frequently and extremely fast in a non-stationary way~\cite{gama2010knowledge}. Furthermore, in most applications, the labels that are required to accurately measure the model performance are not immediately available, which can lead to a late detection of anomalous events in a model life cycle, such as a hot sale item (where the model could be made less strict to reduce false alarms), a fraud attack (where the model could be made more strict to block more fraud attempts) or a data issue (e.g., an API changes and suddenly some important data fields are not available, so a software fix needs to be developed and installed). To tackle this problem, we propose an Automatic Machine Learning (\textit{autoML}) system for model monitoring in data streams. It was designed to detect sudden changes in behaviour occurring in relatively short time scales, from a few hours to a few days, in a situation where labels are not immediately available. The system, SAMM (Streaming system for Automatic Model Monitoring), is able to detect sudden changes, but also to provide relevant information to explain them, a very important feature to understand what caused them especially if there is a client in the loop.

In recent years the use of ML models in production environments became a widespread practice. Very frequently an application requires more than one model, several machines with different environments, receives data from several types of devices, in different geographical locations, just to name a few of the complexities. This wide scope for unexpected behaviour or sudden changes makes the task of {\em model monitoring} extremely challenging if done by humans, and it creates an urgent demand for systems like SAMM.

In the data streams literature, these sudden changes are recognized as concept drift~\cite{gama2014survey}. Gama et al. defined concept drift as a change in the joint distribution between a set of input variables and a target variable~\cite{gama2010knowledge}. ML algorithms for streaming data must be able to cope and adapt to concept drift. Most available methods in the literature tackle this problem by assuming that the labels are immediately (or \textit{almost} immediately) available after prediction. This enables them to use the loss of the predictive model to detect concept drift. However in many use cases, where labels are collected with several weeks of delay, this is not realistic. SAMM uses the stream of scores produced by the model to detect local changes in their distribution.
\begin{figure}[b]
	\hspace{-6pt}\includegraphics[scale=0.37]{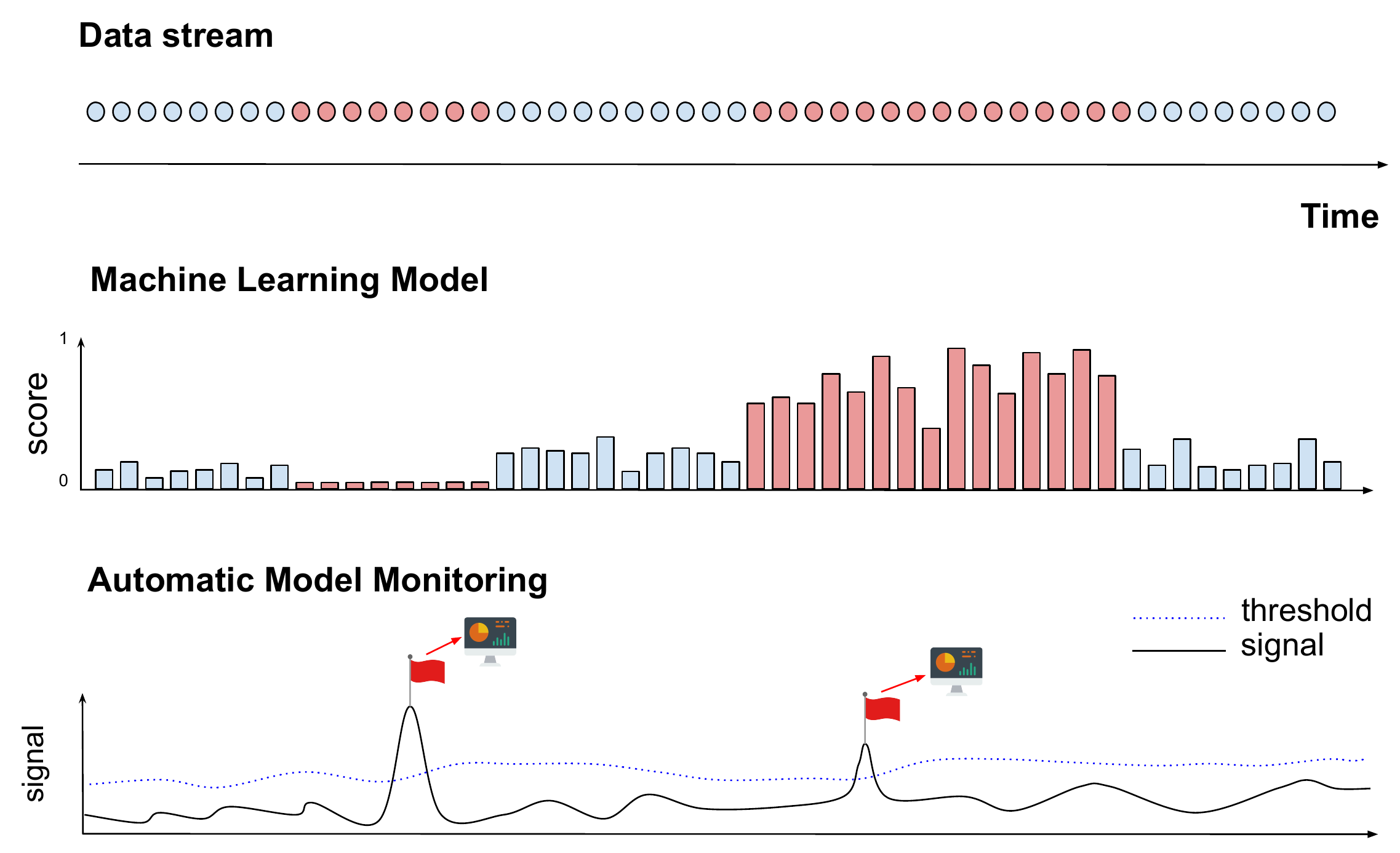}
	\caption{\label{fig:scheme}Schematic overview of SAMM.}
\end{figure}
Figure~\ref{fig:scheme} presents a schematic overview of SAMM. Let us assume, for simplicity, that the model is responsible for detecting the positive examples of a binary classification task. The top plot in Figure~\ref{fig:scheme} represents the evolution of a data stream, in which new patterns (coloured in light red) may suddenly appear and impact the ML model performance. Let us also assume, for example, that there is a period of normal behaviour (coloured in light blue) followed by periods of anomalous behaviour (in light red). In the middle plot of Figure~\ref{fig:scheme} we represent the time series of model scores produced by the model. Assume that the first period of anomalous behaviour is a bot attack in a fraud detection scenario. Furthermore, assume that the model could not detect this attack well, as seen by the low risk scores in the middle plot of Figure~\ref{fig:scheme}. In our system this shift in the scores distribution is captured by a signal $S$ (bottom plot in Figure~\ref{fig:scheme}) that provides a measure of similarity between the model scores distribution in a target window $T$ (most recent examples) and in a reference window $R$ (see Section~\ref{subsec:SignalComputation} for details). This first example consists of a fraud attack that the model was not able to block. The same applies for fraud attacks that the model is able to correctly block, as depicted in the second anomalous period in the middle plot. This is also useful information if certain fraud attacks, that are known to the model, may overload the production system.

In the bottom plot of Figure~\ref{fig:scheme}, we represent how $S$ evolves over time as the data stream of model scores changes. If $S$ is larger than a given threshold $\tau$ (see Section~\ref{subsec:AdaptiveThreshold} for details) an alarm is triggered. For each alarm that is triggered, we then launch an auxiliary process to provide an \textit{explanation}, where another ML model is trained for the task of finding the pattern that best distinguishes the examples in $T$ from the ones in $R$. The output score and the feature importance of that auxiliary ML model is then used to summarize the characteristics of the alarm (see Section~\ref{subsec:AlarmReport}).

In summary, our contributions with SAMM are the following:
\begin{itemize}
	\item In Section~\ref{subsec:SignalComputation}, we introduce a method to compute the signal and the threshold used to detect concept drift in an unsupervised way.
	\item In Section~\ref{subsec:AdaptiveThreshold}, we propose SPEAR, a constant time and memory streaming algorithm for percentiles estimation ($O(n)$ in the fixed number $n$ of bins).
	\item In Section~\ref{subsec:AlarmReport}, we introduce a method to produce explanation reports for the drifts. We also introduce a method to remove time correlated features from the model used for the reports.
	\item In Section~\ref{sec:Experiments}, we present experimental results with five real world datasets, in which we show that our method is able to detect anomalous events (validated by users).
\end{itemize}


\section{Related Work}
\label{sec:RelatedWork}

Recently, there has been a growing interest in researching \textit{autoML} methods~\cite{golovin2017google,cortes2017adanet,liang2019evolutionary} -- the sub-field of ML research that focuses on developing methods with the aim of automating different stages of a ML workflow. This interest is largely due to: 1) an increase in the amount of stored data, and 2) the shortage of qualified experts to develop ML systems. Early research on \textit{autoML} has mainly focused on model selection and hyperparameter tuning~\cite{kotthoff2017auto,feurer2015efficient,mendoza2016towards}. Recently, however, there has been some effort to extend such methods to stages such as data cleaning and pre-processing~\cite{sutton2018data}, feature engineering and selection~\cite{kaul2017autolearn,nargesian2017learning}, and more recently, model monitoring~\cite{madrid2018towards}. Our contributions in this paper focus on the latter.

In the data streams literature, the problem of model monitoring has been tackled mostly as concept drift detection~\cite{gama2014survey}. The two most frequently used methods to detect concept drift are ADWIN~\cite{bifet2007learning} and the Page-Hinkley test~\cite{gama2013evaluating}. Both methods take the loss of the predictive model as one of the inputs (e.g., the error rate). Therefore, they assume that the labels are immediately (or almost immediately) available to compute the loss. This is not always the case in real world applications of ML systems where, depending on the application, labels can take weeks to be available. For this reason, and because of the growing number of industrial applications of ML systems, research on unsupervised concept drift detection has been receiving more attention.

In our literature review, we found that the papers by {\v{Z}}liobaite~\cite{vzliobaite2010change} and dos Reis et al.~\cite{dos2016fast} present approaches that relate the most with ours. In the former, the method consists of measuring the difference in the distributions of the classifier output in two consecutive sliding windows of the same size, by applying a statistical test such as Kolmogorov-Smirnov, Wilcoxon rank sum or t-test. In the latter, the authors propose an incremental Kolmogorov-Smirnov test for faster processing time and in their experiments, one of the sliding windows (the oldest) is not consecutive to the most recent sliding window, but fixed. Though in both papers the results were favourable to the detection of drifts in an unsupervised fashion, the evaluation was either performed by introducing artificial drifts (for example by corrupting feature values) or by focusing on detection rates based on the target variable of the predictive model (which excludes drifts that are unrelated to the target variable). In contrast, in our study, we focus our evaluation on real datasets with no artificial injection of drifts, and we rely instead on user feedback to assess the reliability of the alarms that our system triggers.

Sethi and Kantardzic~\cite{sethi2017reliable} propose MD3, a method that measures the density of points in the uncertainty regions produced by margin-bounded classifiers, such as SVM. According to the authors, since the method is designed to focus on the classifier performance, MD3 produces a reduced number of false alarms, enabling the triggering of drifts only when they are most likely to affect the classifier performance. Shujian et al.~\cite{yu2018request} propose a hierarchical method that includes two layers of hypothesis tests. Surprisingly, although using significantly fewer labels, their methods outperforms supervised drift detectors like DDM~\cite{gama2004learning}.

More recently, there has been a growth in attention for methods that merge the automation of model monitoring and drift detection systems. Madrid et al.~\cite{madrid2018towards} extended the popular ML tool kit \textit{scikit-learn} to include drift detection methods that allow to automatically detect when models should be updated. However, the method requires labels. Contrastingly, Ghanta et al.~\cite{ghanta2019ml} describe a system that does not require labels since it only monitors features distributions to detect drifts. This implies that their monitoring is strictly univariate, which can lead to an increase in false positives.

\section{Method}
\label{sec:Method}

In this section we present the various components of SAMM, namely the signal (Section~\ref{subsec:SignalComputation}) that is used to trigger an alarm if it grows larger than the threshold (Section~\ref{subsec:AdaptiveThreshold}), and the report that is generated for each alarm (Section~\ref{subsec:AlarmReport}).

\subsection{Signal Computation}
\label{subsec:SignalComputation}

In our method, the signal value $S$ is computed for each incoming event. The signal consists of a measure of similarity between the model scores histogram in a reference window $R$ and the model scores histogram in a fixed-size target window $T$. The $T$ window contains the last $n_T$ events collected. As for $R$, it contains events in a reference period, which is prior to the target period and it consists of the $n_R$ events immediately before $T$ (fixed-size window), or of the events in a window with a fixed time duration ending at the oldest event in $T$ (fixed-time window). In our study we prefer to use fixed-size windows, though we can apply it to fixed-time windows as well. This gives us a better control of our estimators, since it fixes the dependency of the variance on the sample size. In contrast, for time-based windows, when comparing signal values for two different events, we would be comparing signal values computed with two different sample sizes. This would be particularly worrying for the $T$ window, which is typically smaller. The $R$ window size is chosen to be some multiple of the $T$ window size (e.g., 5 times larger).

As mentioned before, we designed our system with the goal of detecting sudden changes in behaviour occurring in relatively short time scales, from a few hours to a few days, in a situation where labels are not immediately available.  This configuration is particularly well suited to this goal because it provides a comparison between the $T$ window and the most recent events preceding it.
If, however, the data contains strong seasonality, say certain types of events tend to occur more at certain times of the day, it may induce repetitive daily alarms that will not be so useful.
These seasonality effects can in principle be removed by adapting the $R$ window. For example, if the periodicity is daily, then the $R$ window could be composed of several replica windows in preceding days where each replica window lasts for a period homologous to the $T$ window, but on the previous day, two days ago, three days ago, and so on. The datasets we have analysed in this study did not suffer from this issue, so a more extensive discussion of homologous windows will be left to future work.

Regarding window sizes, we choose the size of the $T$ window in units of the average number of events in some period (e.g., one hour, half a day or one day). Our preliminary studies showed that the sizes of $R$ and $T$ affect directly the amount of noise present in the signal. Very short windows tend to generate noisy signals, which result in more false alarms; very large windows can make the signal insensitive to small changes in the distribution of scores. As a rule of thumb, our preliminary studies showed that 3 and 0.5 times the average number of daily events are good default values for the sizes of $R$ and $T$, respectively, for the type of datasets we analysed. We use these parameter values to compute $S$ in the remainder of the paper, and defer a more thorough sensitivity study to a future work.

The signal $S$ is defined using a measure $M$ of similarity between histograms such that $S = M(R,T)$. In this study we will present results using the Jensen-Shannon Divergence (JSD)~\cite{jsd_article}, though other measures are possible. In preliminary studies we also analysed the Kolmogorov-Smirnov, Kuiper and Anderson-Darling test statistics as distance measures~\cite{massey1951kolmogorov,kuiper1960tests,anderson1954test}. Except for differences in the noise level (higher in some cases), we did not find these alternative signals to be substantially different. On the other hand, the JSD has some appealing information theoretical properties. Namely, it is a measure of the mutual information between the random variable generated by a binary mixture model of the two distributions and the corresponding binary indicator variable. Furthermore, similarly to the other measures, it is bounded and symmetric. When the distributions are the same, it goes to zero, whereas when they have disjoint domains it goes to $\log 2$ (or $1$ if entropy is measured in shannon units). Finally, the JSD is also trivially computed for multi-dimensional distributions, making it a good solution to compute $S$ in multi-class model monitoring use cases. However, in this paper, we only perform experiments for binary classification problems.

\subsection{Threshold}
\label{subsec:AdaptiveThreshold}

In this Section we present a method to compute a threshold, for the signal, above which an alarm is triggered. In a stationary environment, a simple procedure to define outlier values for the signal is to flag all values that fall in the upper tail of the distribution computed with the whole series (e.g., above the $95$th percentile). To compute the threshold using such a definition, we need a reliable and flexible method to estimate percentiles in streaming scenarios. Most methods in the literature resort to sampling previously observed instances and keeping them in memory (e.g., Q-digest or, more recently, the T-digest algorithm~\cite{2019arXiv190204023D_Dunning} -- see also the review by Luo et al.~\cite{Luo2016}). Though sampling can provide high memory compression and high accuracy estimates, it often requires managing clusters of samples including sorting operations. In our study, we do not need high accuracy estimates, thus we adopt a lighter approach with a fixed number of bins updated all at once, with a single linear pass, which can then be used to estimate any percentile through interpolation. Our approach amounts to a stochastic approximation of the cumulative distribution function. Earlier studies following similar density estimation strategies are~\cite{tierney1983,Chen2000IQE347090.347195,Naumov2007ExponentiallyWS,Hammer2019}. Our approach differs in that we use a simple percentiles update strategy anchored on the principle of restoring the invariant (on each new event) that the average count per bin is the same for all bins.

We design an algorithm and related data structure, named SPEAR (Streaming Percentiles EstimAtoR), with the following properties:
\begin{itemize}
	\item {\em Space efficiency:} we only save a fixed size $O(n)$ object with the positions of $n+1$ percentile estimates $\mathbf{P} \equiv \left[P_0,P_1,\ldots, P_n\right]$, where $P_0$ and $P_n$ provide estimates of the lower/upper range of the domain of the distribution, respectively.
	\item {\em Time efficiency:} the time complexity for each incoming event is $O(n)$, so that on any new event all percentiles are updated in a single pass over the percentiles object.
	\item {\em Streaming implementation:} each event is processed only once and the new estimate $\mathbf{P}$ only depends on the last estimate.
\end{itemize}

We will show empirically, with real datasets, that the algorithm works well enough to be used with SAMM. A formal study of the convergence properties and estimation efficiency of the algorithm is out of the scope of this paper and it will be left to future work.

\begin{algorithm}
	\caption{SPEAR -- Consumes a stream of values in real time\label{algo:SPEAR}.}
	\begin{algorithmic}[1]
		\State stream = \Call{newStream}{ }
		\State $\mathbf{P} \gets []$  \Comment{Initialise global empty list of approximate percentiles}
		\State $C=0$   \Comment{Initialise count variable}
		\While{stream.\Call{notClosed}{ }}
			\State $X =$ stream.\Call{getValue}{ } \Comment{Get last value received}
		    \State $C \gets C + 1$ \Comment{Update count}
			\If{$C\leq n$} \Comment{Initialise $\mathbf{P}$ using first $n+1$ values}
				\State $\mathbf{P} \gets$ $\mathbf{P}$.\Call{InsertSorted}{$X$}
			\Else \Comment{Update $\mathbf{P}$ otherwise}
				\State $\mathbf{P} \gets$ \Call{UpdatePercentiles}{$\mathbf{P}$, $X$, $C$}
			\EndIf
		\EndWhile

		\State
		\Function{UpdatePercentiles}{$\mathbf{P}$, $X$, $C$}
		\State $c_{\rm per\_bin} \gets C/n$ \Comment{Counts per bin before update}
		\State $c_{\rm target}\gets (C + 1)/n$ \Comment{Target counts per bin after update}
		\State
		\State $c_{\rm this}\gets c_{\rm per\_bin}$ \Comment{Set count at current bin (${\rm bin}_1$)}
		\If{$X < P_0$} \Comment{$X$ to the left of ${\rm bin}_1$}
		\State $P_0\gets X$ \Comment{Left-expand ${\rm bin}_1$ (move $P_0$ left)}
		\EndIf
		\If{$X < P_1$} \Comment{$X$ in ${\rm bin}_1$ or to the left}
		\State $c_{\rm this}\gets c_{\rm this} + 1$ \Comment{Increase count to add $X$ to ${\rm bin}_1$}
		\EndIf
		\State
		\For{$i \gets 1,\ldots ,n-1$} \Comment{Move internal bin walls}
		\State $\delta c \gets c_{\rm target} - c_{\rm this}$ \Comment{Deficit count at ${\rm bin}_i$}
		\If{$\delta c > 0$} \Comment{${\rm bin}_i$ smaller than target}
		\If{$X < P_{i+1}$} \Comment{$X$ in ${\rm bin}_{i+1}$}
		\State$\rho_{\rm next} = \frac{1 + c_{\rm per\_bin}}{P_{i+1} - P_{i}}$ \Comment{Computes ${\rm bin}_{i+1}$'s density}
		\Else
		\State $\rho_{\rm next} = \frac{c_{\rm per\_bin}}{P_{i+1} - P_{i}}$ \Comment{Computes ${\rm bin}_{i+1}$'s density}
		\EndIf
		\State $P_i \gets P_i + \delta c / \rho_{\rm next}$ \Comment{Right-expand ${\rm bin}_i$}
		\State $c_{\rm this} \gets \rho_{\rm next} (P_{i+1} - P_{i})$ \Comment{Saves ${\rm bin}_{i+1}$ new count}
		\Else
		\State $\rho_{\rm this} = \frac{c_{\rm this}}{P_{i} - P_{i-1}}$ \Comment{Density at ${\rm bin}_i$}
		\State $P_i \gets P_i + \delta c / \rho_{\rm this}$ \Comment{Left-expand ${\rm bin}_{i+1}$}
		\State $c_{\rm this} \gets c_{\rm per\_bin} - \delta c $ \Comment{Saves ${\rm bin}_{i+1}$ new count}
		\EndIf
		\EndFor
		\State
		\If{$X > P_{n}$} $P_n \gets X$ \Comment{Right-expand ${\rm bin}_n$}
		\EndIf
		\State \textbf{return} $\mathbf{P}$
		\EndFunction
	\end{algorithmic}
\end{algorithm}

In Algorithm~\ref{algo:SPEAR} the pseudo-code is presented. The first 10 lines represent the sequential consumption of the values as they stream into the system. Inside the {\em while} loop, for the first $n+1$ values that stream in, the values are inserted into the global list $\mathbf{P}$ in sorted order, to initialise an estimate of the $n+1$ percentile positions. For simplicity, here we assume that the values streaming in are almost surely unique. If that's not the case, which is often true in practice if round off occurs or if the distribution of values is discrete, the initialisation step can be modified to include some numerical noise, so that all initial percentile position values are unique. The percentile position estimates are later updated when more events stream in, so the impact of this initialisation should fade away as more data is collected. The subsequent values for $C>n$, are processed on line 10. For each incoming event the percentile position estimates $\mathbf{P}$ are updated taking into account the incoming value $X$ and the current total count $C$.

\begin{figure}
	\includegraphics[width=7cm]{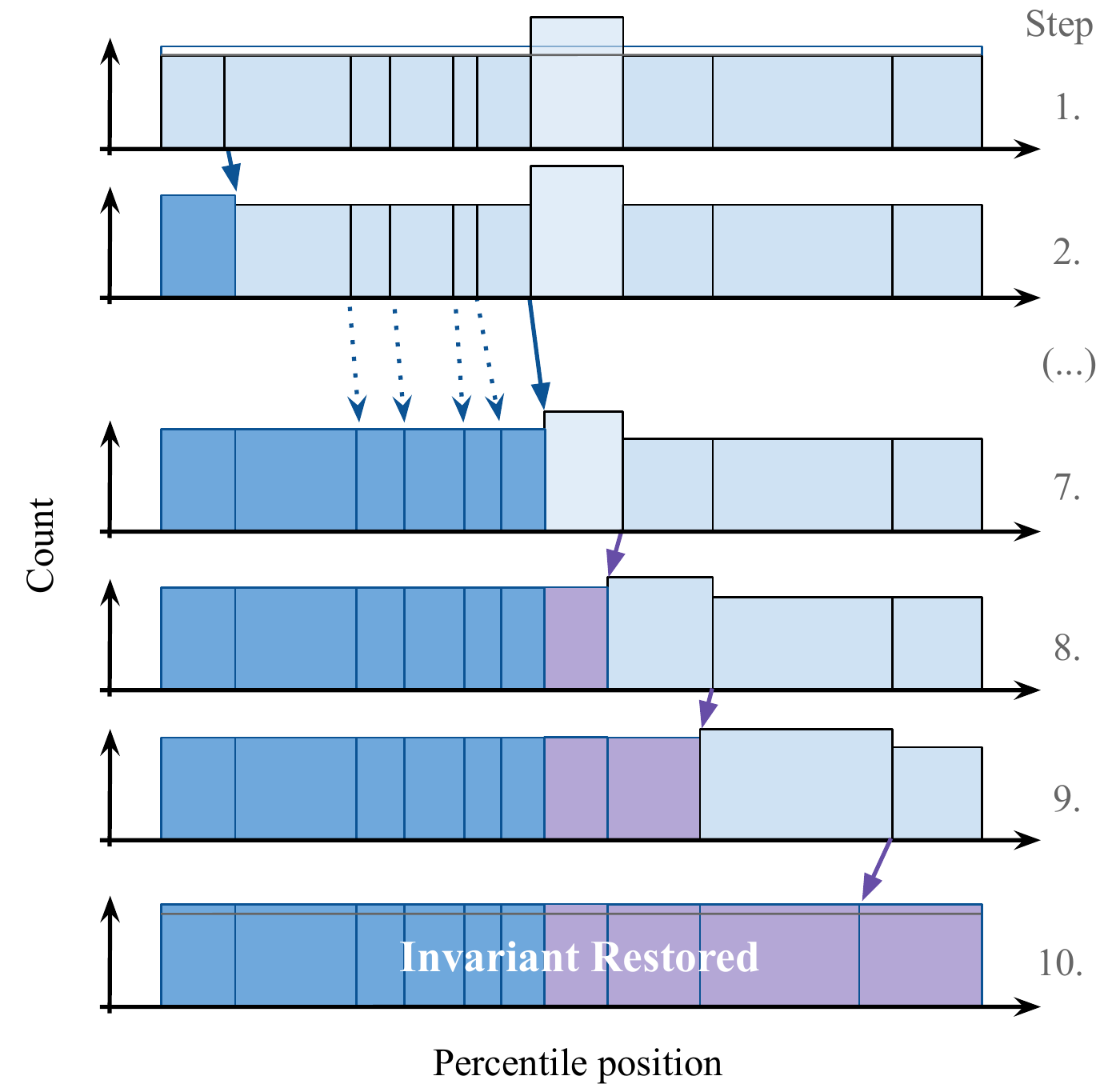}
	\caption{\label{fig:UpdatePercentiles} One call of \textsc{UpdatePercentiles}. 1) Adds event to bin 7 ($c_{\rm target}$: blue horizontal line ). 2) to 7) Right-expand Bin walls on the left. 8) to 10): Left-expand other bins walls.}
\end{figure}

The bulk of the work of the algorithm is done by the \textsc{UpdatePercentiles} function. The algorithm works by maintaining the invariant that the estimated number of counts in each bin is the same for all bins. The new value $X$ streaming in, will belong to some ${\rm bin}_j$ for which the count will increase by 1. This breaks the invariant. To restore the invariant, in \textsc{UpdatePercentiles}, we first compute the new target count-per-bin, which is given by the mean number of events per bin after adding the new event. Then we loop over all bins from left to right. We expand, to the right, bins that have a deficit count relative to the target value. This ``eats'' a portion of the next bin based on its density. When we encounter $X$'s bin, the bins start having an excess count so we contract them (or, equivalently expand the next bin to the left), ``shedding'' away a portion to the next bin, based on the current bin density. Fig.~\ref{fig:UpdatePercentiles} illustrates this procedure.

This algorithm is asymmetric, because the \textsc{UpdatePercentiles} function operates from left to right. This will create a directional bias in the estimate. A straightforward way to correct this bias is to also apply the update from right to left and average out the two:
\begin{eqnarray}
\mathbf{P} \hspace{-4pt}&\gets& \hspace{-6pt}\tfrac{1}{2}\left[\textsc{UpdatePercentiles}\left(\mathbf{P}, X_{n + k}, C\right) +\right.\\
&&\hspace{-6pt}\left.\textsc{Reverse}(\textsc{UpdatePercentiles}\left(\textsc{Reverse}(-\mathbf{P}), -X_{n + k}, C\right)\right] \; ,\nonumber
\end{eqnarray}
where \textsc{Reverse} returns the elements of a list in inverted order. A greedier option that avoids duplicating the amount of work, is to choose between a left-right or right-left pass on each new incoming event either in an alternate way or with equal probability (to avoid reintroducing bias if the stream contains unfavourable correlations).

\subsection{Alarm Report}
\label{subsec:AlarmReport}

After an alarm is triggered, a natural question for the user is \textit{``What caused the alarm?''}. Knowledge of the signal, alone, does not provide any information on the characteristics of the subset of events in the $T$ window that caused the alarm.

In this section, we introduce a strategy to guide the user in analysing the alarm. An automatic report is generated by comparing the events in the $T$ and $R$ windows through a measure of dissimilarity. The goal is to rank the $T$ window events according to how likely they are to explain the alarm. Even though we only use the model score\footnote{This can be seen as an aggregated view of the event focused on the target.} to compute the signal, other features of the events may provide further useful information. Therefore we employ a strategy that leverages the power of a ML model that uses both the features and the model score. On each alarm, we create a new target binary label with value $1$ for events in $T$ and value $0$ for events in $R$ and train an auxiliary ML model to learn how to separate events in the two windows. We use a Gradient Boosted Decision Trees (GBDT) model. This allows us to: 1) obtain an {\em alarm score} that can be used to rank events in $T$ (higher score $\Rightarrow$ closer to the top), and 2) directly obtain a measure of feature importance that deals well with correlated features~\cite{xu2014gradient}. The latter provides a way of ranking the features themselves. The hyperparameters of the GBDT model were fixed to 50 trees with a maximum depth of 5, for all reports in all experiments. All other hyperparameters were kept with the default value of the scikit-learn implementation~\cite{pedregosa2011scikit}.

\begin{figure}
	\includegraphics[width=5.2cm]{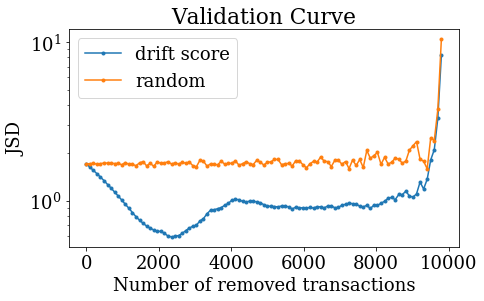}
	\caption{\label{fig:ValidationPlot}Example validation plot for an alarm report.}
\end{figure}

To test the robustness of the ranking provided by the GBDT model, we formulated an independent validation method. Since the goal of the ranking is to push to the top the events that are responsible for distorting the distribution of model scores in the target window, we expect that removing events from the top of the list will suppress the signal. Therefore, we define a validation curve (see Figure~\ref{fig:ValidationPlot}) where each point is the value of the signal using $R$ as reference, but $T$ with the top $k$ events removed. For comparison, we define a curve where, for each point, we randomly remove $k$ events from $T$. The latter should not be able to lower the signal value if the alarm is in fact a false positive. In that case the blue curve (removal by drift score) should be similar or above the yellow one (random).

In summary, our alarm report contains the following:
\begin{itemize}
\item Window information with start and end timestamps,
\item Truncated feature importance ranking list (e.g., top 10),
\item Validation curve to observe how well the ranking can lower the signal,
\item A table of the top $N$ (e.g., 100) events that explain the alarm. This may contain some extra fields, that are informative, selected according to domain knowledge (e.g., emails, addresses, card identifiers) and it contains the feature values used by the GBDT model (with columns ordered from left to right according to the feature importance ranking).
\end{itemize}

\subsubsection*{Removal of time correlated features}

One potential issue of the ML model approach presented above is that some features may be correlated with time or, similarly, the index that defines the order of the events. Then, due to the sequential nature of the window configuration ($T$ comes after $R$), those features will allow the model to very easily learn how to separate the $T$ window events from the $R$ windows events using that time information (instead of learning the differences in the distributions of features between the two windows). To address this problem we apply a pre-processing method in a {\em burn in period}\footnote{We define the {\em burn in period} as a set of initial events in the data stream that is used for initialisation. This includes filling up windows (so that the signal can be computed) and the computation of time correlated features discussed in this section.} to detect features that correlate with time. Those features are then excluded from the training of the GBDT model for the alarm report.

\begin{figure}
\hspace{-5pt}\includegraphics[height=3.1cm]{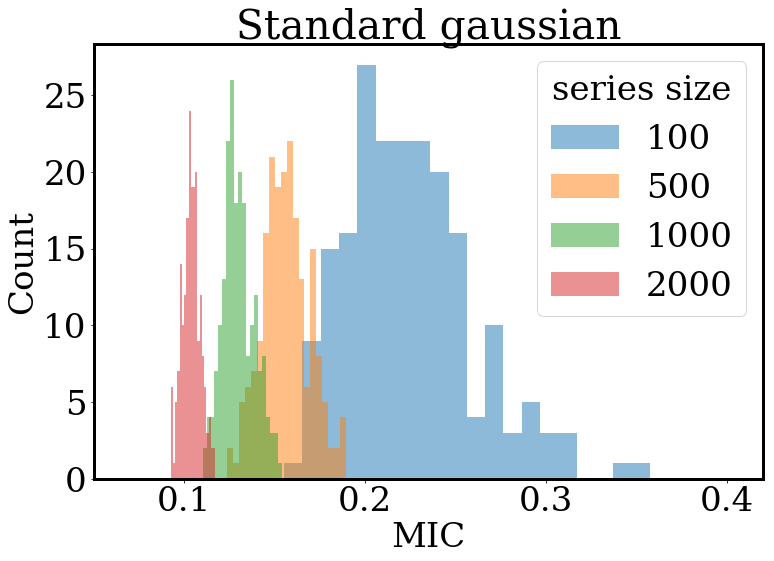}\includegraphics[height=3.1cm]{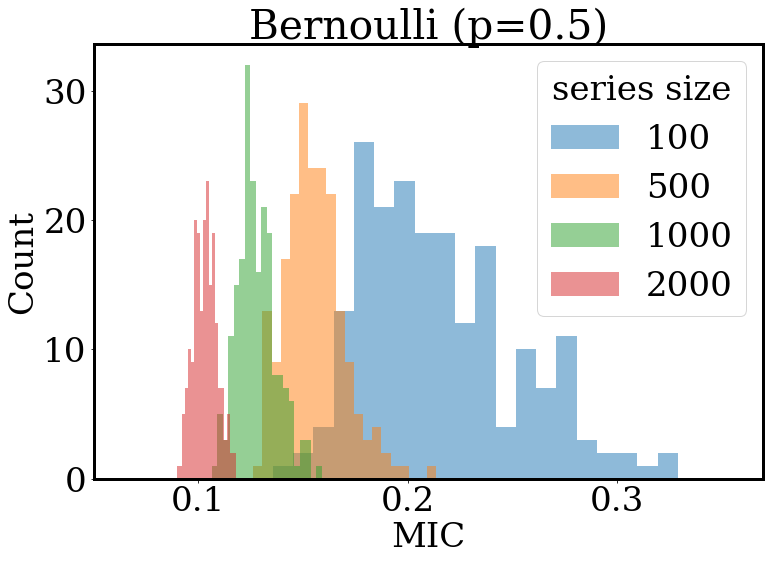} \\
\caption{\label{fig:MIC} MIC distributions for some source distributions. }
\end{figure}

To formulate the method, consider a time series:
 \begin{equation}
\left[(t_0, X_0), \ldots,(t_i,X_i),\ldots, (t_N,X_N)\right]
\end{equation}
We want to know whether there is a correlation between the ordered set of timestamps (or index values) $\mathbf{T} = \left[t_0,\ldots, t_i,\ldots,t_N\right] $ and the feature values $\mathbf{X} = \left[X_0,\ldots, X_i,\ldots,X_N\right] $. We use a measure of correlation that is sensitive to non-linear relations, the Maximal Information Coefficient (MIC)~\cite{Reshef1518}, which is bounded in the interval $[0,1]$ ($MIC=1$ corresponds to a perfect correlation). Given $\mathbf{X}$ and $\mathbf{T}$ and $MIC(\mathbf{X}, \mathbf{T}) \neq 0$ we also need a measure of significance (under the null hypothesis $H_0$ that $MIC(\mathbf{X},\mathbf{T}) = 0$). This is done by setting an $\alpha$-level on the probability of observing the MIC value due to random chance. The probability distribution of MIC depends on the distribution of values that generates the given series, as well as on the size of the series. A direct way to estimate it numerically for specific $(distribution, size)$ pairs is with Monte Carlo (MC) simulation, by sampling (several times) a series of fixed size from a fixed distribution, and by computing the MIC value for each sample. In Figure~\ref{fig:MIC} we show some distributions with $200$ MC samples each, for gaussian numerical features (left panel) and a categorical bernoulli feature (right panel). This shows that, as expected, the variance is reduced as we increase the series size. We have also checked that the MIC distribution only varies mildly with the source distribution used to generate the series.

Computing the full MIC distribution with simulation, to determine the p-value, is unnecessarily heavy. Instead we ask a simpler related question: \textit{``How many $M$ samples of MIC do we need to observe under $H_0$, to have seen a value at least as large as $MIC_\alpha$ with probability at least $p$?''}. This is given by
\begin{eqnarray}
P\left(\max\left(MIC_1,\ldots,MIC_M\right)\geq MIC_\alpha\right) &=& 1-(1-\alpha)^M \geq p \nonumber\\
\Rightarrow M & \geq & \dfrac{\log(1-p)}{\log(1-\alpha)}
\end{eqnarray}
For simplicity we set $p=1-\alpha$. If $\alpha=0.05$, this means that we need $M\simeq 60$ to have a $95\%$ probability to obtain one MIC value (or more) in the $5\%$ upper tail of the distribution.

This discussion assumes that we have access to the distribution of values that has generated the data. Since we work with streams of data with sizes above the thousands of instances we assume that the series $\mathbf{X}$ itself provides a good estimate of its distribution of values. Thus, for simplicity, we generate the $M\simeq 60$ values for MIC by simply shuffling the series randomly $M$ times and computing the corresponding MIC values for each shuffle. The maximum observed value serves as a threshold for the feature $\mathbf{X}$,  which is removed if $MIC(\mathbf{X})$ is larger. In the real datasets we analysed, for computational time efficiency reasons, we used series of size $1000$ covering the $R$ plus $T$ window, with points uniformly spaced in the {\em burn in} period.

\section{Experiments}
\label{sec:Experiments}
\begin{table}[!b] \centering
	\begin{tabular}{cccccccc}
		\\[-1.8ex]\hline
		\hline \\[-1.8ex]
		Dataset & Features & Days & Transactions & Transactions per day\\
		\hline \\[-1.8ex]
		$A1$ & $213$ & $212$ & $\num[group-separator={,}]{1046482}$ & $\num[group-separator={,}]{4936}$ \\
		$A2$ & $213$ & $212$ & $\num[group-separator={,}]{2667548}$ & $\num[group-separator={,}]{12583}$  \\
		$A3$ & $213$ & $212$ & $\num[group-separator={,}]{4945509}$ & $\num[group-separator={,}]{23328}$ \\
		$B1$ & $279$ & $229$ & $\num[group-separator={,}]{4401807}$ & $\num[group-separator={,}]{19221}$ \\
		$B2$ & $279$ & $229$ & $\num[group-separator={,}]{9229013}$ & $\num[group-separator={,}]{40301}$  \\
		\hline \\[-1.8ex]
	\end{tabular}
	\caption{\label{table:datasets}  Summary statistics for each dataset-region.}
\end{table}
In this section, we provide results from experiments with real world datasets.
We used two datasets in the fraud detection domain with more than 20 million online transactions (see summary statistics in Table~\ref{table:datasets}). The various fields were collected in a production environment. Each dataset is comprised of several regions (3 for dataset $A$ and 2 for dataset $B$) and for each region there is one model responsible for scoring its transactions. Dataset $A$ consists of credit card payments from an on-demand mobility company. Dataset $B$ consists of online payments from an e-commerce merchant.

To validate the alarm reports, in principle, we would need labels for the alarms themselves. The prior existence of those labels relies on records that users of the system may (or may not) have produced for past anomalous events. This poses an extra difficulty to evaluate SAMM. Thus, our evaluation strategy was to actively ask for user feedback, from domain experts, on 100 reports generated by the system. In Section~\ref{subsec:results}, we will describe in detail the experimental setup and the results we obtained. Before discussing them, we present an example of the output obtained for the signal and threshold in Section~\ref{subsec:SignaAndThresholdExperiments}, and we discuss the validation of alarm reports in Section~\ref{subsec:ValidationOfReports}.

\subsection{Signal and threshold}
\label{subsec:SignaAndThresholdExperiments}

\begin{figure}
	\includegraphics[width=8.5cm]{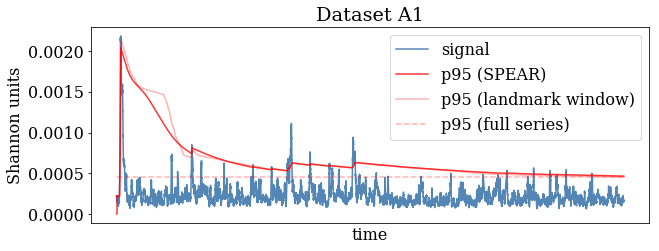}
	\caption{\label{fig:signal} Signal and thresholds for datasets \textit{A1}. }
\end{figure}

In Figure~\ref{fig:signal} we show an example of the signal and threshold computed for dataset~\textit{A1}. The signal was computed as described in Section~\ref{subsec:SignalComputation} and is coloured in blue. The SPEAR algorithm threshold is coloured in red. For comparison purposes, we also show a light red line where the 95th percentile was computed (exactly) with a landmark window\footnote{This window contains all the events since the beginning of the dataset.}. SPEAR is, in fact, estimating well the 95th percentile computed with the landmark window, especially at later times (after accumulating enough examples). Finally, the dotted red line represents the 95th percentile computed with the full series. As expected, both the SPEAR and the landmark window lines converge to the 95th percentile of the full series.

For the experiments with dataset \textit{A}, we used the 95th percentile estimated with the SPEAR algorithm.
For dataset \textit{B}, we decreased the threshold to the 85th percentile. The choice to lower the threshold for this use case was based on domain knowledge we were provided with, hinting that more anomalous events were to be expected in the models from dataset \textit{B}. We present a discussion of this choice and how it affected the results in Section~\ref{sec:Discussion}.

\subsection{Validation of the reports}
\label{subsec:ValidationOfReports}

Since collecting drift labels in production environments is an extremely difficult task, we faced the need to perform independent validations of the drifts that were detected. One of such validation strategies was already mentioned in Section~\ref{subsec:AlarmReport} in Figure~\ref{fig:ValidationPlot} for an alarm report.
\begin{figure}[b]
		\includegraphics[width=5.2cm]{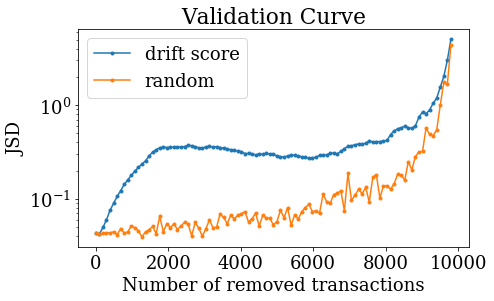}
	\caption{\label{fig:ValidationPlotValley}Example validation plot for a valley.}
\end{figure}
In contrast, Figure~\ref{fig:ValidationPlotValley} shows an example of a validation plot for a valley. A valley, in this experiment, is a data point in $S$ where the JSD is close to 0, for which we expect the transactions in $R$ and $T$ to be very similar. The alarm reports for valleys should be confusing and difficult to interpret for the data scientists, since it corresponds to a time frame where the data stream should be homogeneous. As expected, since the examples in $R$ and those in $T$ are similar, removing the top ranked examples from $T$ does not lead to a decrease in the JSD value. This is one of the most important mechanisms that we have found to validate if a given alarm is a true positive or a false positive.

\begin{figure}
	\includegraphics[width=4cm]{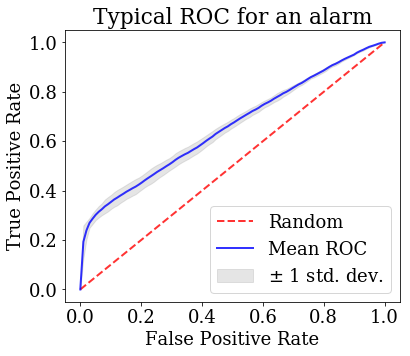}\includegraphics[width=4cm]{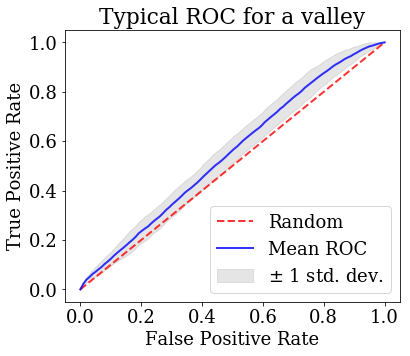}
	\caption{\label{fig:rocs} Examples of ROC curves generated with 5-fold cross validation for an alarm (left) and a valley (right).}
\end{figure}

Another source of validation that can be used, consists of performing an evaluation with k-fold cross validation, by re-fitting GBDT models, with the same hyperparameters as before, on each fold. This allows us to estimate the ROC curve. We run this procedure for each alarm and compute the average AUC. Although the examples in our use case are time ordered, for this particular purpose, k-fold cross validation is still a suitable technique, since we are not interested in an unbiased estimate of performance for a model that is able to generalize into the future. Furthermore, observe that we have previously removed the time correlated features, so time dependencies should be mild. Figure~\ref{fig:rocs} shows examples of ROC curves computed from a 5-fold cross validation procedure. Typically, for alarms, the ROC curve behaves similar to the one in the left panel, where the model is able to learn how to distinguish the examples in $R$ from the ones in $T$. For valleys, the performance is usually very close to random, therefore the ROC curve is similar to the random line. This indicates that, as expected for valleys, the model is not able to learn any pattern to separate the examples.

\subsection{Experimental Results}
\label{subsec:results}

 We designed an experiment where we had two data scientists available for each dataset-region. We generated 100 alarm reports (20 per dataset-regions). Each data scientist received a list of 10 alarm reports to analyse, with the structure defined as in Section~\ref{subsec:AlarmReport}. This means that each alarm report contained the top 100 transactions to be observed. For each alarm, the data scientists were asked to score, from 1 to 5, the following questions:

\begin{itemize}
	\item \textbf{Q1.} Please provide a score from 1 to 5 reflecting how confident you are that this report corresponds to a true alarm  (1: totally sure it is false, 5: totally sure it is true)
	\item \textbf{Q2.} Please provide a score from 1 to 5 reflecting how clearly you can see a pattern in the transactions of the report provided (1: no pattern at all, 5: very clear pattern)
	\item \textbf{Q3.} After looking at the validation plot, please provide a new answer to Q1, regarding this new information\footnote{The validation plot was hidden from the user for Q1 and Q2.}.
\end{itemize}

To evaluate the outcome of these experiments, we mixed the reports based on alarms with reports based on valleys. The distribution of alarms and valleys was set randomly by a Bernoulli variable with $p=0.5$ (equal probability to select an alarm or a valley). This selected 52 reports based on alarms and 48 reports based on valleys. Our hypothesis was that the reports based on alarms would have higher scores in the three questions asked to each data scientist.

The statistical analysis of the results was done using the Mann-Whitney U test with the Holm-Bonferroni correction for multiple comparisons~\cite{mann1947test,holm1979simple}. In all tests, the $\alpha$-level for the null hypothesis, that there is no difference between alarms and valleys, was $0.05$.

\begin{figure*}[!tbp]
	\centering
	\begin{subfigure}{.33\textwidth}
		\includegraphics[scale=0.38]{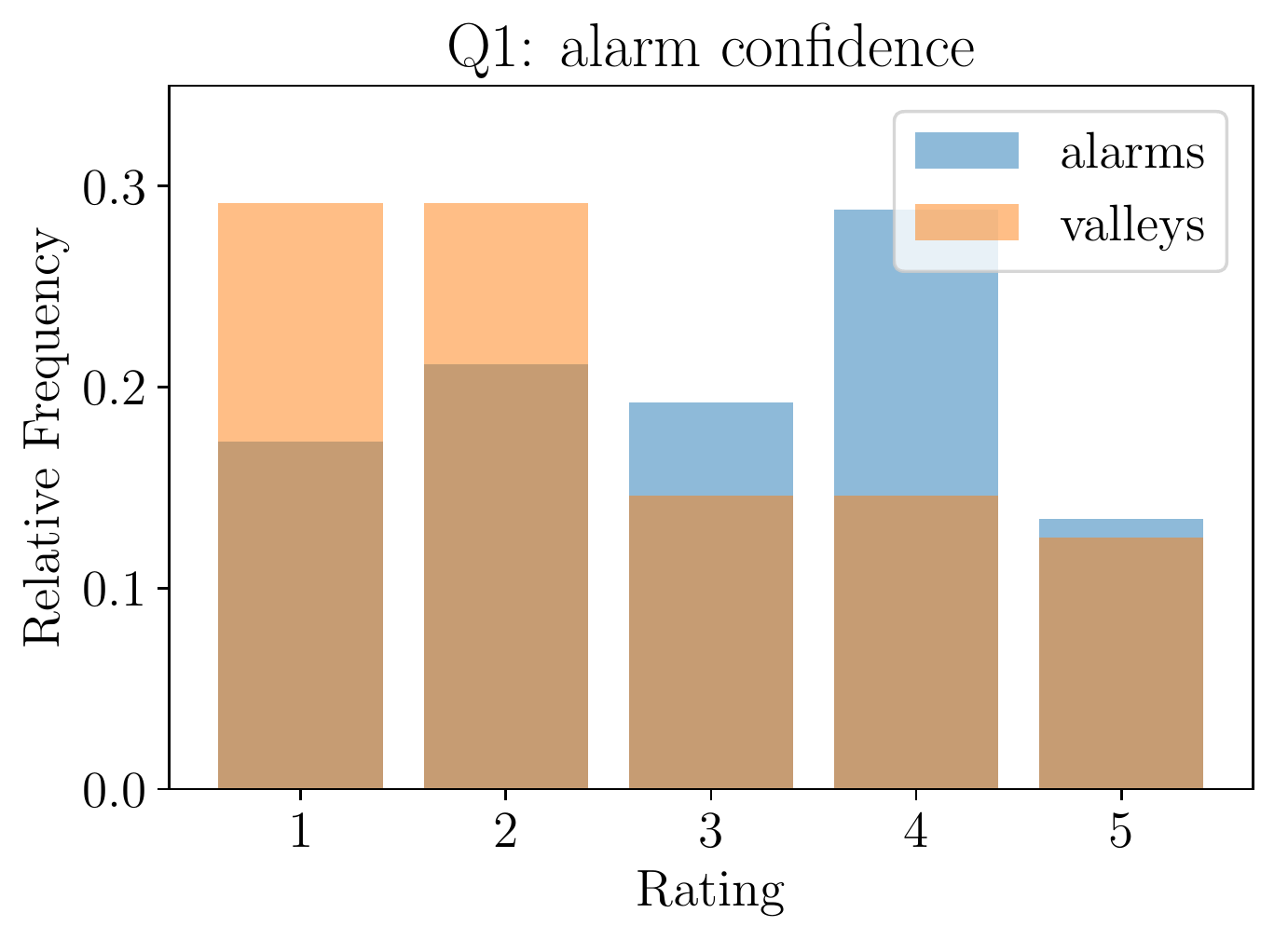}
	\end{subfigure}%
	\begin{subfigure}{.33\textwidth}
		\includegraphics[scale=0.38]{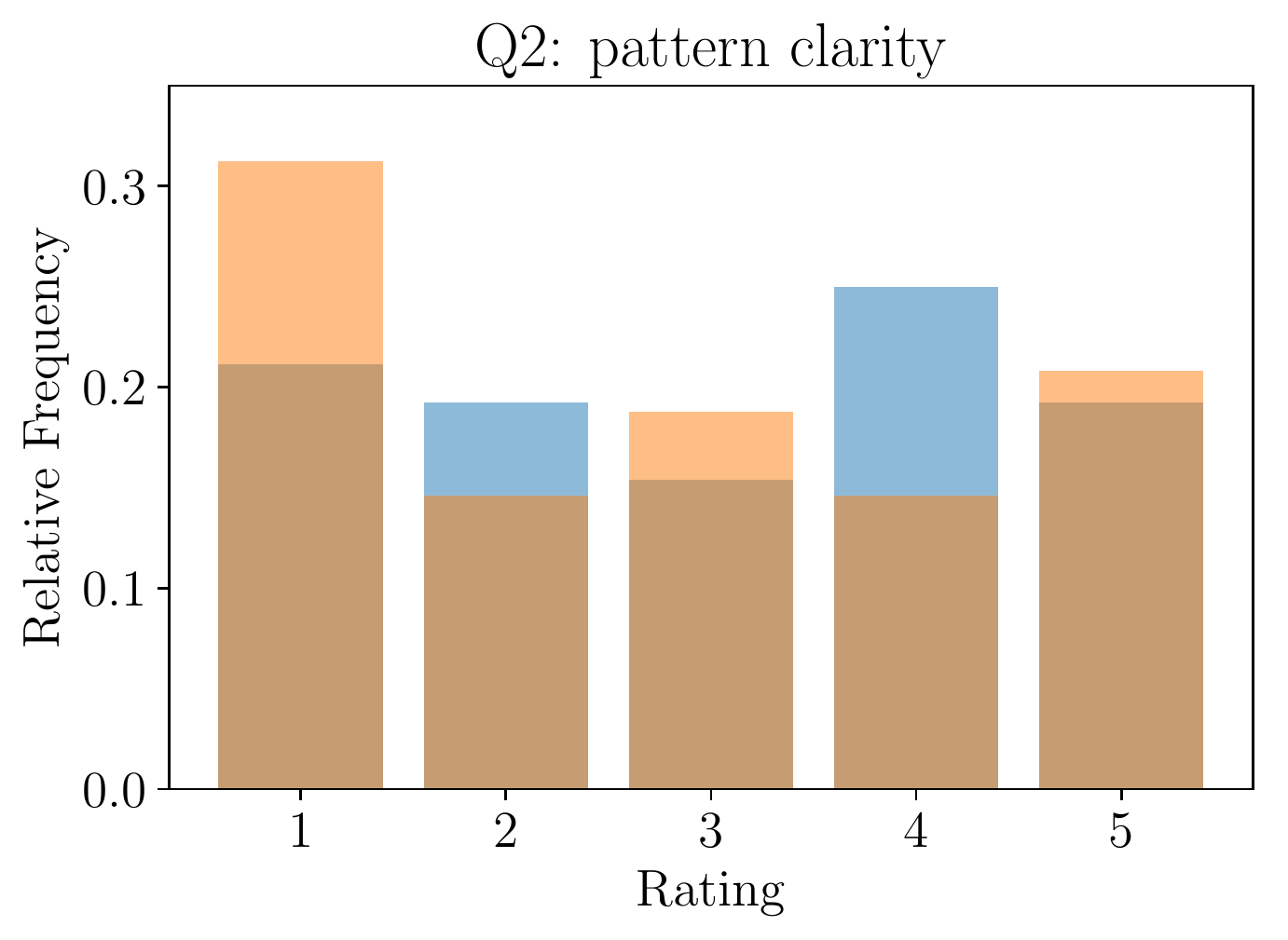}
	\end{subfigure}%
	\begin{subfigure}{.33\textwidth}
		\includegraphics[scale=0.38]{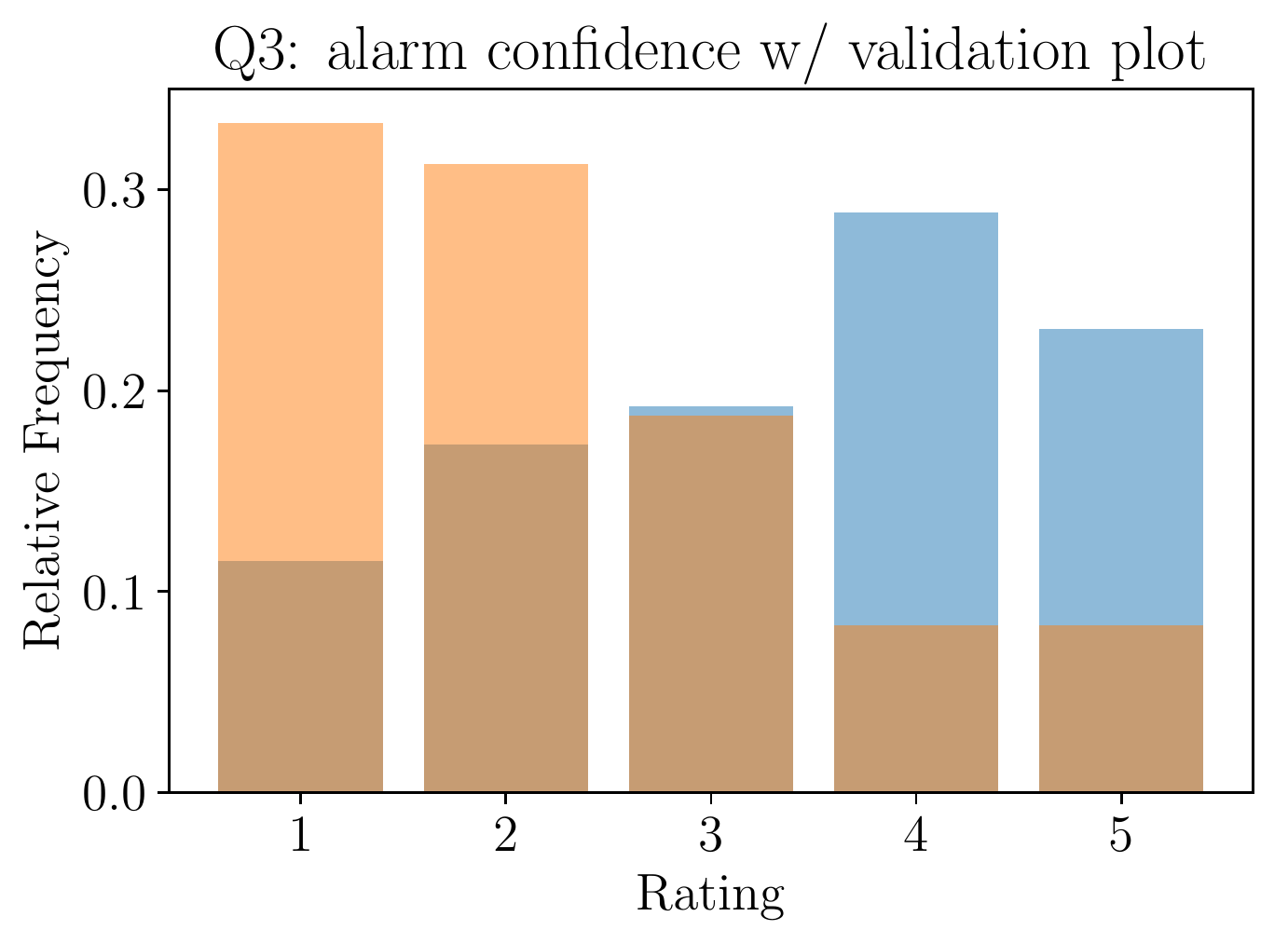}
	\end{subfigure}
	\caption{\label{fig:agg_plots} Bar plots with score distribution for alarms and valleys collected from the experiments.}
\end{figure*}

The distribution of ratings for the three questions asked in the experiments is shown in Figure~\ref{fig:agg_plots}. The distributions for alarms are represented in blue bars whereas for valleys they are in orange. The output of the statistical tests, to determine if ratings for alarms are significantly larger than for valleys, is summarized in Table~\ref{fig:results}. For \textit{Q1} and \textit{Q3}, there is a clear separation between alarms and valleys: the former tend to have higher ratings in both questions. The separation is even more notorious in \textit{Q3}. Regarding statistical significance, for the aggregated view (with all the reports from datasets \textit{A} and \textit{B}) both \textit{Q1} and \textit{Q3} present a p-value smaller than the defined $\alpha$-level. This shows that the difference observed in Figure~\ref{fig:agg_plots} is statistical significant. For \textit{Q3}, this still holds in the breakdown by dataset. However, for \textit{Q1}, the p-value for dataset \textit{B} is 0.370, which is above the $\alpha$-level. For that particular dataset, the results were, overall, not so good, possibly indicating that many of the reports based on alarms could in fact be false positives. We discuss this in more detail in Section~\ref{sec:Discussion}.

Regarding \textit{Q2}, the aggregated view and the breakdown for dataset \textit{B} show a p-value above $\alpha$. This indicates that the differences visible in the \textit{Q2} plot of Figure~\ref{fig:agg_plots} are not statistically significant. However, the p-value for dataset \textit{A} is 0.037, below the $\alpha$-level. One of the possible reasons for this result in the aggregation, is that the question itself is more subjective.
In this question, the evaluators were asked how clearly could they see the pattern in the transactions displayed in the report. This is intrinsically more difficult because it requires a deeper investigation. Moreover, feedback collected from the participants afterwards, showed that some of them could not spend additional time analysing the reports. This may have resulted in outcomes where the participants were not able to detect a pattern in the reports even if they rated the alarm highly in \textit{Q1}.
Thus, the subjectiveness of the question demands for more time from the participants to properly answer it.
We plan to take this into consideration in future work and guarantee that all participants allocate the same amount of time to analyse the reports.

\begin{table}[!tbp] \centering
	\begin{tabular}{@{\extracolsep{5pt}} rrrr} 
		\\[-1.8ex]\hline 
		\hline \\[-1.8ex] 
		& Q1 & Q2 & Q3 \\ 
		\hline \\[-1.8ex] 
		Aggregated & $\textbf{0.037}$ & $0.220$ & $\textbf{5.6e-5}$ \\ 
		Dataset A & $\textbf{0.037}$ & $\textbf{0.037}$ & $\textbf{0.003}$ \\ 
		Dataset B & $0.370$ & $0.822$ & $\textbf{0.006}$ \\ 
		\hline \\[-1.8ex] 
	\end{tabular}
	\caption{\label{fig:results} P-values for the various Mann-Whitney U tests (with Holm-Bonferroni correction for comparisons by dataset). For all tests, the $\alpha$-level was set at 0.05. Values in bold represent tests where the p-value was smaller than $\alpha$.}
\end{table}

\section{Discussion}
\label{sec:Discussion}

The results presented in Section~\ref{subsec:results} indicate that, according to user feedback, the alarm reports contain more true positive alarms than false positive alarms, relative to periods of low signal (\textit{Q1}). Furthermore \textit{Q3} also shows that the validation plot adds extra confidence to the user. For \textit{Q2}, the alarm reports were not, overall, significantly useful in helping to identify clearly the pattern responsible for the alarm.

However, focusing on the breakdowns per dataset (suitably corrected for multiple testing), we have seen that, in fact, the p-values are consistently much worse for dataset \textit{B} and that, in fact, statistical significance is attained for all questions for dataset \textit{A}.

\begin{figure}[b]
	\includegraphics[width=8.5cm]{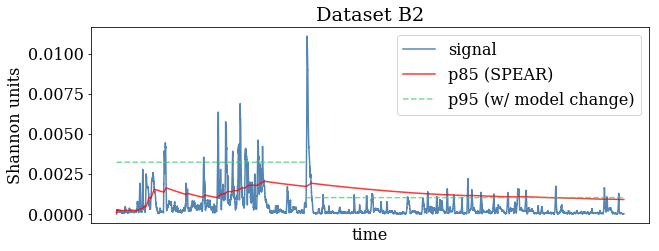}
	\caption{\label{fig:signal_adaptive} Signal and threshold for dataset \textit{B2}.}
\end{figure}
We now discuss a possible reason for the results of dataset \textit{B}.
Figure~\ref{fig:signal_adaptive} shows the signal and thresholds for dataset \textit{B2} as an example. In that plot we can see a very high peak that reaches above the 0.01 shannon units mark. This was due to a new model deployment that, as expected, led to a transition to a new regime for the distribution of model scores. The dotted green line represents the 95th percentile before this event (using all data points until the deployment date) and after this event (using all data points after the deployment date). We can observe that there is a clear regime change in the series. In the experimental setup for dataset~\textit{B}, we made the choice of lowering the threshold (in fact this corresponds, approximately, to the 95th percentile after the new model deployment). We suspect that this choice increased the amount of false positive alarms in the pre-deployment period, thus contributing to more confusing alarm reports (see red line for \textsc{SPEAR} in Figure~\ref{fig:signal_adaptive}, where the signal oscillates much more above the threshold).

Since new model deployments are quite common events in Data Science projects, this shows the importance of using a method that can deal with such abrupt regime changes, to be able to adapt the threshold accordingly. Ideally, such method would also be parameter free (or almost parameter free). We plan to address this issue in the future with an adaptive version of the SPEAR algorithm.

Regarding the evaluation strategy for our system, we chose to rely on direct user feedback on generated alarm reports. This provides valuable information, because it tries to simulate some aspects of the interaction a user would have with the system in a real production environment. In Section~\ref{sec:Experiments}, we have also mentioned that another possible strategy would be to try to collect past records of abrupt events that users have kept in production. Though this is more difficult to collect systematically, it is extremely valuable information, because users typically have time to cross check those events over days or weeks of work. Furthermore, these records will tend to contain the most important alarms (i.e., those that have a larger business impact), which means that the labelling will be more solid. We plan to aggregate available records in a future study to enrich our evaluation strategy.

\section{Conclusions}
\label{sec:Conclusions}

This paper proposes SAMM, an \textit{autoML} system for automatic monitoring of ML models for data streams. The two combined characteristics that distinguish SAMM from the state-of-the-art are that it detects concept drift in an unsupervised way and it suggests an explanation for the drift in the form of an alarm report. Although there are some methods already proposed in the literature for unsupervised concept drift detection, to the best of our knowledge, our system is the first proposal attempting to explain why a drift occurred.

We evaluated SAMM in five real world fraud detection datasets, for which we generated 100 alarms reports and resorted to human feedback to score three questions: 1) how confident the user was that the report corresponded to a true alarm 2) how clearly the user was able to see a pattern in the report, and 3) if the user wanted to change the answer to the first question after seeing a validation plot. The human feedback was provided by data scientists that worked daily with the datasets. Our results showed higher scores for alarm reports for questions 1 and 3 (relative to reports for valleys). For both questions, the differences were statistically significant. Regarding question 2, the differences were statistically significant only for one of the datasets.

In future work, we plan to extend our method to be able to adapt the threshold to regime changes. One possibility that we are investigating is the use of forgetting factors, a technique that is often used in data streams~\cite{gama2013evaluating}. Another interesting topic would be to investigate how data streams with high seasonality would affect the signal that we compute, which was not an issue for the datasets used in our experiments.
Such effect is to be expected in some datasets, so we plan to address it in the future. Finally, we also plan to study the effect of the contents presented in the alarm reports from an UX perspective. Information overload can be a problem and potentially impact the usefulness of such reports. Ideally, an alarm report would only present the strictly necessary information for the user be able to quickly understand the drift.


%
\bibliographystyle{ACM-Reference-Format}
\bibliography{biblio}

%




\end{document}